\documentclass{ar-1col-S2O}
\usepackage[numbers]{natbib}
\usepackage{url}
\usepackage{float}

\jname{Annu. Rev. Control Robot. Auton. Syst.}
\jvol{9:7.1-7.27}
\jyear{2026}
\doi{10.1146/annurev-control-032724-014418}

\begin{document}

\markboth{Milford \arblcirc{} Fischer}{Going Places}

\title{Going Places: Place Recognition in Artificial and Natural Systems}

\author{Michael Milford and Tobias Fischer
\affil{QUT Centre for Robotics, School of Electrical Engineering and Robotics, Queensland University of Technology, Brisbane, Queensland, Australia, 4000; email: michael.milford@qut.edu.au, tobias.fischer@qut.edu.au}}

\begin{abstract}
Place recognition---the ability to identify previously visited locations---is critical for both biological navigation and autonomous systems. This review synthesizes findings from robotic systems, animal studies, and human research to explore how different systems encode and recall place. We examine the computational and representational strategies employed across artificial systems, animals, and humans, highlighting convergent solutions such as topological mapping, cue integration, and memory management. Animal systems reveal evolved mechanisms for multimodal navigation and environmental adaptation, while human studies provide unique insights into semantic place concepts, cultural influences, and introspective capabilities. Artificial systems showcase scalable architectures and data-driven models. We propose a unifying set of concepts by which to consider and develop place recognition mechanisms and identify key challenges such as generalization, robustness, and environmental variability. This review aims to foster innovations in artificial localization by connecting future developments in artificial place recognition systems to insights from both animal navigation research and human spatial cognition studies.
\end{abstract}

\begin{keywords}
place recognition, loop closure, simultaneous localization and mapping, SLAM, scene representations, robotics, hippocampus, entorhinal cortex, bioinspiration, nature, animals, insects, humans, navigation, neuroscience, biology
\end{keywords}
\maketitle

\section{INTRODUCTION}
Place recognition---the process of an animal, person or robot recognizing a familiar location in the world---has attracted significant attention across multiple disciplines. In animals, this capability has evolved over millions of years through sophisticated neural mechanisms: hippocampal place cells fire at specific spatial locations~\cite{okeefe1976}, entorhinal grid cells provide spatial coordinates through hexagonal firing patterns~\cite{hafting2005}, while diverse species demonstrate remarkable navigation—from desert ants using celestial cues and visual panoramas~\cite{wehner2003} to migratory birds returning to precise breeding sites across hemispheric distances~\cite{wiltschko2005}. Humans extend these biological foundations with unique cognitive abilities, recognizing places not only through sensory perception but also through semantic meaning, emotional associations, and cultural context—enabling us to identify familiar locations from descriptions, memories, or even fictional narratives~\cite{tulving2002}. In artificial systems, place recognition underpins core robotics functions such as localization, mapping, and long-term autonomy, developing into a mature field that, while sometimes inspired by biological principles, often diverges significantly in implementation to optimize for computational efficiency and metric accuracy.

As research has grown in the area, so too has a rich landscape of surveys and reviews that reflect the field's evolution and diversification. The computational place recognition literature has matured from foundational reviews to specialized deep learning and multi-modal approaches, while neuroscience has established fundamental principles of spatial cognition across species. However, existing surveys remain largely domain-specific, creating an opportunity for a single article covering and synthesizing all domains, aiming to identify shared principles, points of significant differentiation, and mutual learning opportunities. Those outcomes are the aim of this review article.

The computational place recognition survey landscape shows clear temporal and methodological evolution. Lowry et al.~\cite{Lowry2016VisualSurvey} established the foundational framework for visual place recognition (VPR), introducing core terminology and systematic coverage of traditional handcrafted methods during the pre-deep learning era. With the deep learning revolution came a new generation of surveys. Zhang et al.~\cite{Zhang2021VisualPerspective} and Masone \& Caputo~\cite{Masone2021ARecognition} provided comprehensive treatments of CNN-based approaches, with Zhang et al.~offering a primarily methodological focus on deep learning architectures and Masone \& Caputo bridging traditional and modern approaches through systematic pipeline analysis. Tsintotas et al.~\cite{Tsintotas2022TheDetection} focused specifically on loop closure detection within Simultaneous Localization and Mapping (SLAM) systems, emphasizing practical deployment challenges. Recent surveys reflect increasing specialization and practical concerns, including research best practice, and improvement of and unification of terminology. Garg et al.~\cite{Garg2021WhereRecognition} offered conceptual innovation by redefining VPR within a broader Spatial Artificial Intelligence framework. Schubert et al.~\cite{Schubert2024VisualTutorial} introduced the first dedicated VPR tutorial, providing hands-on implementation guidance with modern deep learning methods. Yin et al.~\cite{Yin2025GeneralAutonomy} expanded beyond visual-only approaches with comprehensive multi-modal coverage (visual, LiDAR, radar) for real-world autonomy. While much focus in place recognition has been on the visual (camera) sensing modality, maturing sensors and techniques have also led to modality-specific surveys addressing non-visual sensing approaches. Zhang et al.~\cite{Zhang2025LiDAR-BasedSurvey} and Luo et al.~\cite{Luo20243DSurvey} covered LiDAR and 3D point cloud methods respectively. 

In the biological and neuroscience literature, the concept of place recognition has unfolded over the past fifty years following the pivotal discovery of place cells in the early 1970s ~\cite{okeefe1976}, a neuron type that was observed to fire only when the animal, initially a rodent, was located at a specific place. In these fields, the focus has primarily been on discovering and characterizing a number of different spatially-responsive cells, investigated through a combination of experimental paradigms, typically in lab-like environments, and computational neuroscience modelling. In parallel, but mostly on a separate research track, researchers have investigated the navigational capabilities of many organisms, but these studies, especially the ones in larger, more naturalistic environments, have mostly been behavioral in nature, looking at migration journeys, exploration and foraging activity but often without detailed neuronal recordings, with some notable exceptions in organisms like bats \cite{eliav2025fragmented}. 

Despite this rich literature, significant gaps exist in integrated, cross-domain coverage of both computational approaches and the natural systems that have evolved to perform this task. Computational surveys typically include only brief mentions of biological inspiration (usually limited to citations of place or grid cells), missing the wealth of potential insights from diverse animal navigation strategies—from the snapshot matching of insects to the magnetic map sense of sea turtles. Similarly, they rarely consider human spatial cognition, which uniquely combines metric, topological, and semantic representations in ways that could inform more flexible artificial systems. To be clear, the proposal here is not that the developers of artificial systems should slavishly mimic every aspect of natural systems. Rather, we propose that there are many detailed aspects of how natural systems perform place recognition and related tasks that are interesting to consider when designing artificial systems, and that much of the current literature does not jointly cover these issues in sufficient depth to enable this to occur. Conversely, neuroscience work rarely considers implications for artificial systems or points of overlap. To take but one example: whilst active range sensing is rare in the natural kingdom, echolocating animals like bats achieve a sophisticated 3D awareness through biological sonar: a point of particular interest given the focus in computational approaches on both explicit and implicit three-dimensional representations of the structure and layout of the environment. Existing surveys also lack comparative analysis across different intelligent agents (animals, humans, robots) that could reveal universal principles of spatial cognition. The temporal scope varies dramatically -- neuroscience papers often span decades of research while computational surveys often focus on recent -- just the last few years -- technical developments.  

These disconnects and differences provide an opportunity for mutual enrichment, as biological principles could inform more robust computational designs while computational insights could suggest new hypotheses about neural mechanisms. Key commonalities across systems include the fundamental challenge of achieving robust recognition despite environmental variability; the need for efficient memory management; and the integration of multiple information sources. However, critical differences also emerge in how each system addresses these challenges (or perhaps more precisely, how the research field tries to solve these challenges versus how solutions have evolved in nature): animals employ embodied, action-oriented representations shaped by ecological pressures; humans add layers of semantic and cultural meaning; while robots optimize for computational efficiency and metric accuracy. On a more philosophical note, artificial system development is targeted towards both research field norms, like achieving ``state of the art'' performance on benchmarks, and on actual ``deployability'' on commercially-fielded systems like autonomous vehicles. Natural systems have evolved much more slowly as a product of the ecosystems and habitats within which those organisms reside. Understanding these convergences and divergences could inform the development of more adaptive artificial systems while providing new perspectives on biological navigation.

This review touches on these critical gaps by bridging disciplinary domains, and treating biological \emph{and} computational place recognition as complementary investigations of the same fundamental problem. Such a review cannot achieve the same level of detailed depth in all areas that a narrower article can: as such, we focus on the general higher level concepts, with some detailed examples.

The review is structured as follows. There are three main sections, covering place recognition in
\begin{itemize}
    \item robots (and artificial systems in general)
    \item animals (including insects)
    \item humans
\end{itemize}

Each of these sections are themselves split into three repeated subsections addressing key concepts:

\begin{itemize}
    \item what is the nature of what is considered a place in that system?
    \item how are places recognized in that system?
    \item what are the general concepts or insights we can learn from those systems?
\end{itemize}

Sidebars and definitions are provided throughout, focusing on key examples in more detail. Finally, in the Discussion section, we highlight key concepts that we believe should be the focus of future work in place recognition, that combine and highlight both points of difference and commonalities between these different fields.

\section{PLACE RECOGNITION IN ROBOTS}

The importance of, and need for, positioning or localization systems has long been known in robotics~\cite{Leonard1992DirectedNavigation,Dellaert1999MonteRobots}. Place recognition plays an important role within the broader task of positioning and is one of the key enablers of general robot navigation systems, especially in situations where alternative techniques --- like Global Navigation Satellite Systems (GNSS)  --- do not function, including underwater, indoors and underground. Positional knowledge also has widespread applicability beyond just navigation. For example, it can help robots and other autonomous systems make better decisions: augmented reality devices can display location-aware content, interplanetary rovers can repeatedly sample the same locations, and disaster response and monitoring robots can better track and respond to changes in the environment over time.

\begin{marginnote}[]
\entry{Place Recognition}{The ability of a robot to recognize a location it has previously visited, typically by comparing current sensory input to stored representations.}
\end{marginnote}

\subsection{What Is a Place in This System?}
There are competing definitions of what constitutes a place, and these definitions are highly downstream task-specific. For example, when place recognition is used as a loop closure component in SLAM, the matched reference place should be sufficiently close in a geographical sense (pose; i.e., position and orientation, or just position, often using an arbitrary threshold -- for example 25 meters) such that the relative pose between query and reference can be computed. Other researchers argue that the query and reference images should have sufficient visual overlap, which aligns more closely with the image retrieval literature led out of the computer vision field~\cite{Garg2021WhereRecognition}.

\begin{marginnote}[]
\entry{Loop Closure}{The detection that a robot has returned to a previously visited location, used to correct accumulated error in mapping and localization.}
\end{marginnote}

This definitional ambiguity reflects deeper questions about spatial representation in robotics. Some approaches emphasize topological connectivity — places are nodes in a graph where edges represent navigational relationships. Others prioritize metric accuracy, requiring places to maintain precise geometric relationships. Recent work has begun exploring the semantic place concept, where locations are characterized by their functional meaning (e.g., ``kitchen'', ``parking lot'') rather than just visual appearance or geometric properties~\cite{Rosinol20203DHumans}. This final approach connecting more closely to functional meaning is particularly interesting because of the key observation that place recognition -- and localization and navigation more broadly -- is almost always a means, rather than an end. At a functional level, the best ``system for the job'' is the one that best enables the general function of the animal, insect or robot in their habitat or operational ecosystem.

\begin{marginnote}[]
\entry{Semantic Place}{A location defined by its meaning or function (e.g., kitchen, hallway), rather than its appearance or coordinates.}
\end{marginnote}

\subsubsection{Place delineation}
\label{sec:placedelineation}
The most fundamental design choice in developing place recognition systems involves how to segment continuous trajectories (or areas and volumes) into discrete place representations:
\begin{itemize}
    \item \emph{Keyframe-based approaches}, the most widely used paradigm, define places by frames captured at fixed intervals or triggered by motion thresholds. Systems like ORB-SLAM~\cite{Mur-Artal2015ORB-SLAM:System} balance efficiency and coverage, adding keyframes when movement exceeds a distance threshold or visual overlap drops. Adaptive selection based on feature space analysis, rather than geometry alone, can further improve memory efficiency and retrieval performance~\cite{Sheng2019UnsupervisedSLAM,Stathoulopoulos2024WhyRecognition}.
    \item \emph{Segmentation and chunking approaches} acknowledge that places often span extended spatial regions by grouping sequential observations into coherent spatial segments. Sequence-based VPR techniques~\cite{Milford2012SeqSLAM:Nights,Trivigno2022, Garg2021SeqMatchNet:Relocalization,Vysotska2025AdaptiveRecognition} exploit temporal consistency by combining multiple consecutive frames, effectively creating place representations that span extended spatial regions. These approaches must solve the challenging problem of determining appropriate segment boundaries.
    \item \emph{Smooth similarity measures} represent a departure from discrete place definitions, instead treating place recognition as a continuous spatial problem where similarity varies gradually across the environment rather than switching abruptly at place boundaries~\cite{Zaffar2023CoPR:Regression}. Recent approaches weight different spatial regions continuously, producing smooth confidence maps rather than binary place classifications~\cite{Zhu2023R2Recognition}.
    \item \emph{Hierarchical and multi-scale representations} capture places at multiple granularities~\cite{Keetha2021ARecognition}, e.g., ``the kitchen'' (coarse), ``near the refrigerator'' (medium), or ``at the specific viewpoint from last Tuesday'' (fine). Such systems model scale dependency via place trees spanning room-level to viewpoint-specific representations~\cite{Ravichandran2022HierarchicalNetworks}, aiding long-term autonomy (Section~\ref{sec:longtermautonomy}) where both coarse relocalization and precise navigation are needed.
\end{itemize}

\begin{marginnote}[]
\entry{Sequence-Based VPR}{An approach that uses sequences of observations to improve robustness against appearance and viewpoint changes.}
\end{marginnote}

\begin{figure}
\includegraphics[width=4in]{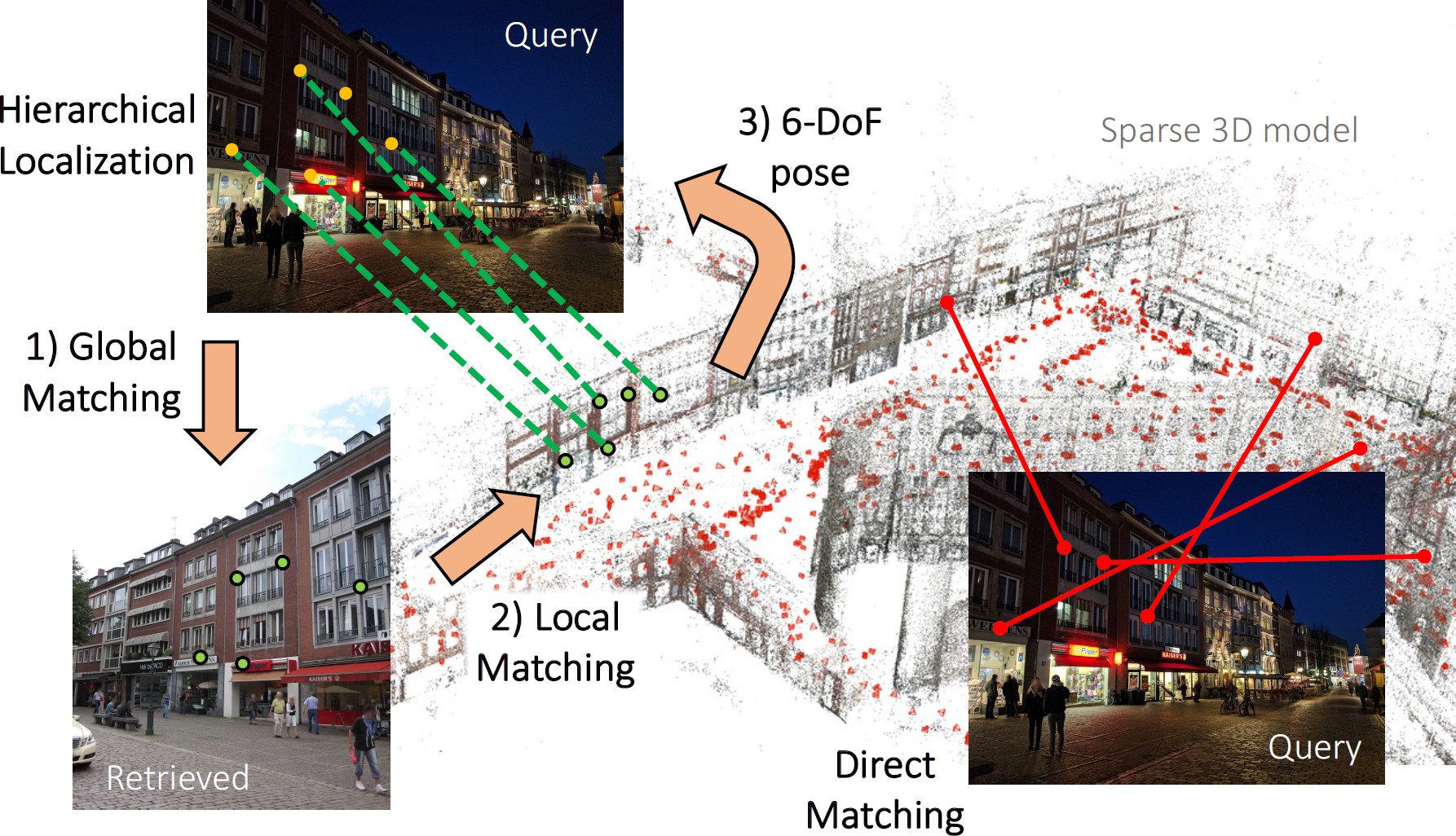}
\caption{From~\cite{sarlin2019coarse}: Hierarchical localization. A global search first retrieves candidate images, which are subsequently matched using powerful local features to estimate an accurate 6-DoF pose. This two-step process is both efficient and robust in challenging situations. Reprinted with permission from the IEEE/CVF Conference on Computer Vision and Pattern Recognition.
}
\label{fig_hierarchical}
\end{figure}

\subsubsection{Downstream tasks}
\label{sec:downstreamtasks}
Place recognition enables a diverse range of functionalities across robotics downstream tasks, each imposing different requirements on the system design~\cite{Schubert2024VisualTutorial}:
\begin{enumerate}
    \item Relocalization addresses the ``kidnapped robot'' problem where a robot must determine its location within a known environment after losing track of its position -- determining whether a robot can recover autonomously from localization failures. This scenario requires robust matching against a pre-built place database and is particularly challenging when the robot's current viewpoint differs significantly from those in the reference map (Figure~\ref{fig_viewpoint_appearance_change}).
    \item Loop closure detection forms a critical component of both online SLAM and offline mapping systems~\cite{Tsintotas2022TheDetection}, where the primary requirement is detecting when a robot has returned to a previously visited location. The success of loop closure depends heavily on the geographic proximity definition of places, as incorrect matches can introduce catastrophic errors into the map. These catastrophic failures occur because loop closure detections are used to enforce global consistency constraints—when a false positive match is accepted, the system attempts to reconcile two locations that are actually far apart by warping the entire map topology. For example, if a robot incorrectly matches a corridor in Building A with a visually similar corridor in Building B, the SLAM backend may force these locations to coincide. Modern SLAM systems often employ hierarchical approaches where VPR provides initial candidates that are then verified through geometric consistency checks to filter out false positives that would damage the map structure (Figure~\ref{fig_hierarchical}).
    \item Topological mapping represents a fundamentally different paradigm where places serve as nodes in navigation graphs, with edges representing traversable connections between locations~\cite{Suomela2024PlaceNav:Recognition}. This application emphasizes connectivity and reachability over precise metric relationships, making it particularly suitable for large-scale navigation where maintaining detailed geometric maps becomes computationally prohibitive. Topological maps are also sometimes all that is \emph{functionally} required for the systems's higher level tasks: long range navigation for example not necessarily needing detailed local metric maps for much of a journey, beyond a ``head in this direction'' type mandate.
    \item Multi-robot map merging presents unique challenges as it requires identifying shared places across maps built by different robots, potentially at different times and under different conditions. This application must handle variations in sensor platforms, viewpoints, and environmental conditions while ensuring that place matches are sufficiently reliable to enable map alignment. A variety of solutions for map sharing have been proposed, including having per-robot front-ends with a shared backend and loop closure detector and direct exchange of compressed maps using mesh networks~\cite{Kottege2024HeterogeneousChallenge,Ebadi2024PresentChallenge}. A common trend in this line of work is to adopt a common sensor payload across heterogeneous robot teams, although this becomes increasingly impractical the more heterogeneous the platforms are, especially with respect to size and payload and power capacity.
\end{enumerate}

\begin{marginnote}[]
\entry{Topological Mapping}{A spatial representation where places are nodes and navigable paths are edges, emphasizing connectivity over precise geometry.}
\end{marginnote}

\subsubsection{Long-term autonomy}
\label{sec:longtermautonomy}
\begin{figure}
\includegraphics[width=3.5in]{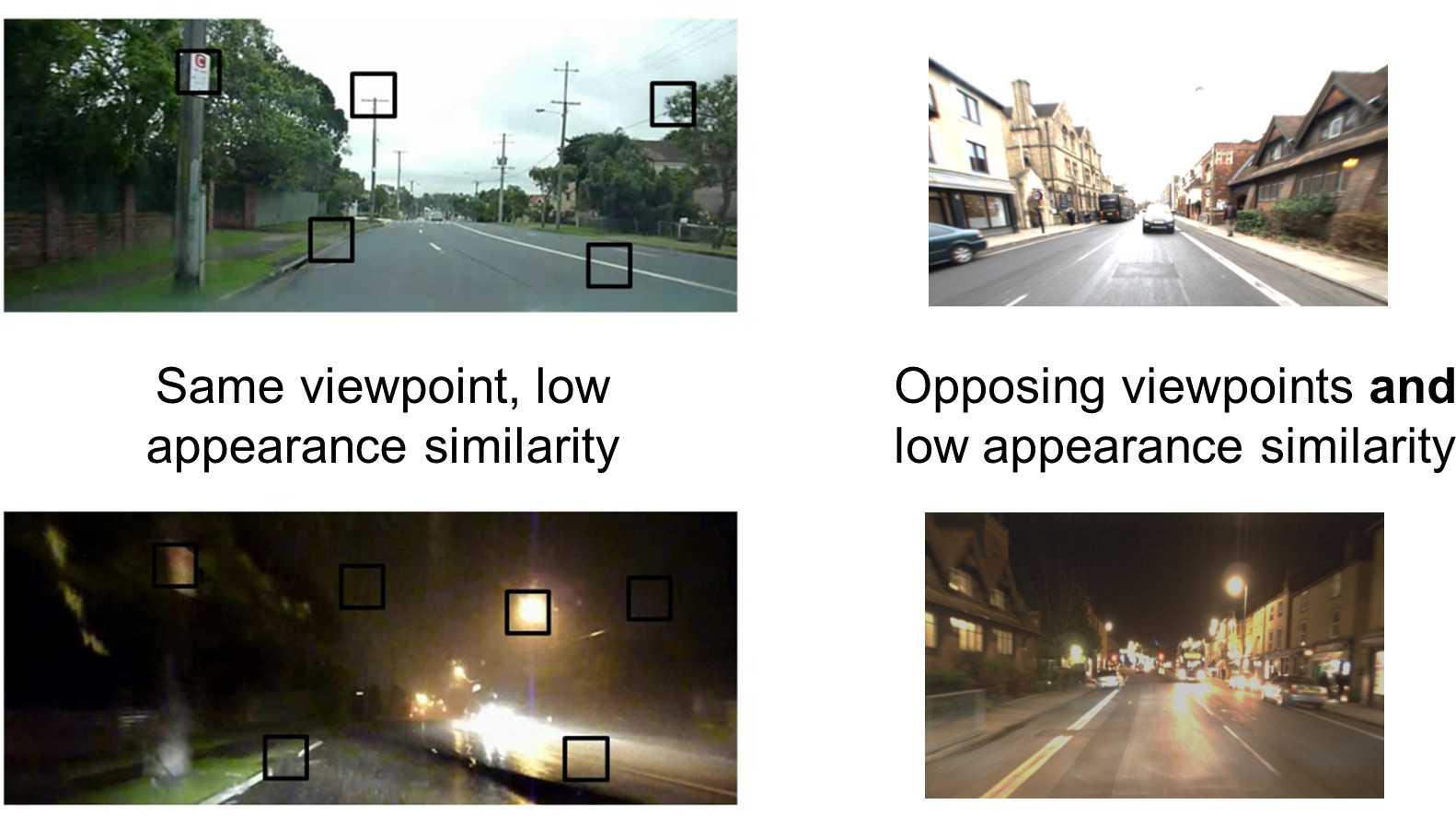}
\caption{\cite{Garg2019SemanticgeometricViews}, \cite{Hausler2021Patch-NetVLAD:Recognition}, \cite{Milford2012SeqSLAM:Nights} and much other work has tackled the key concepts of appearance change and viewpoint change for place recognition, including when both challenges are present simultaneously (right panel).}
\label{fig_viewpoint_appearance_change}
\end{figure}

The aforementioned downstream tasks are required for long-term autonomy, which is a particularly demanding application. It requires place recognition systems to maintain performance as environments change over extended periods -- including handling seasonal variations, construction activities, lighting changes (see Figure~\ref{fig_viewpoint_appearance_change}), and the gradual evolution of indoor spaces~\cite{Doan2019ScalableDriving}. Systems must balance stability (consistently recognizing the same places) with adaptability (accommodating environmental changes) while avoiding catastrophic forgetting of previously learned locations~\cite{Yin2023BioSLAM:Recognition}. A critical capability for long-term deployment is the system's ability to assess the reliability of its own decisions (see ``Introspection in Place Recognition'' sidebar).

One possible solution is to have an ever-growing reference database, where each place is represented with multiple exemplars \cite{churchill2013}. However, this comes at the expense of storage and compute requirements. Instead, a system typically needs to decide when a place has sufficiently changed from its current sample images to warrant addition of a new exemplar. Recent work has shown that it is feasible to merge multiple place exemplars into a single exemplar, while retaining the representational benefits of having multiple exemplars~\cite{Malone2024ARecognition}.

\begin{textbox}[ht]\section{Introspection in Place Recognition}

Would you prefer a place recognition system that works 99.9\% of the time but is unaware when it fails, or one that works 99\% of the time but detects 99\% of its own failures? In safety-critical scenarios like autonomous driving or robotic surgery, this distinction matters greatly: the former may silently fail 0.1\% of the time, while the latter reduces that risk tenfold. Introspection – the ability of a system to assess the quality or reliability of its own decisions – is rapidly becoming a critical capability in modern place recognition pipelines. Unlike traditional systems that produce a single location estimate, introspective systems provide measures of confidence, allowing downstream components to act cautiously, verify outputs, or defer control when uncertainty is high. These abilities mirror adaptive behavior in biological systems, where animals naturally slow down, re-check cues, or switch strategies when navigating uncertain environments.

Recent research in artificial systems is beginning to replicate these introspective functions. One recent approach involves unsupervised quality prediction using internal representations like difference matrices. Carson et al.\ demonstrated that structural cues within these matrices can be used to predict the reliability of single-frame VPR matches, enabling better performance in sequence-based pipelines without the need for ground truth \cite{carson2023}. Other work uses predictive models built on structure-from-motion (SfM) pipelines to anticipate failure by detecting geometric ambiguity or perceptual aliasing before it degrades performance \cite{quteprints257147}. In safety-critical domains, such anticipatory capability is especially valuable, offering time for corrective behavior or system fallback.

Introspective capacity also extends to formal integrity measures. Inspired by techniques from GNSS safety systems, researchers introduced a framework for evaluating whether VPR outputs should be trusted based on internal consistency and statistical cues \cite{carson2022predicting}. These methods enable systems to reject potentially incorrect matches outright, rather than blindly trusting their outputs. Going further, introspective systems are now being used not just to score reliability but to actively improve performance. For example, Claxton et al.\ explored mechanisms for verifying loop closures, discarding low-confidence relocalizations, and even selectively retraining on uncertain matches – enabling systems to modulate reliance on VPR or odometry dynamically, depending on internal confidence estimates \cite{claxton2024improving}.

Together, these efforts mark a fundamental shift from passive to self-aware place recognition. Introspective VPR systems are not just mapping or localizing – they are reasoning about their own perceptual limits and responding accordingly. This capability will be increasingly essential for robots operating in open-world or high-consequence environments. As in biology, knowing when you’re unsure may be just as important as knowing where you are.

\end{textbox}

\subsection{How Is Place Learnt and Recognized?}
\label{sec:learningplaces}
Place recognition systems are typically framed using the retrieval paradigm: previously visited places are encoded and stored as high-dimensional vectors in a reference database~\cite{Schubert2024VisualTutorial}. New incoming sensor data is then matched to this reference database by encoding it and computing similarity to each reference image via cosine or Euclidean distance. This retrieval approach dominates because it naturally handles the open-world nature of robotics applications: robots can recognize places they have previously visited without knowing in advance what those places will be or how many distinct locations they will encounter.

Places are typically represented by global descriptors that are robust to changes in appearance and viewpoint. Global descriptors aggregate information from across the entire image or sensor observation into a single fixed-dimensional vector, in contrast to local descriptors that represent individual keypoints or image patches. The fundamental challenge lies in finding vector embeddings that represent places regardless of capture conditions — whether the place was photographed during daytime or nighttime, in summer or winter, from the perspective of a pedestrian or from a bus, and despite appearance changes that occur over time.

\begin{marginnote}[]
\entry{Global Descriptor}{A (usually) fixed-length vector that captures holistic features of a scene, used for efficient place matching in large databases.}
\end{marginnote}

A common refinement stage employs local descriptors, which are computationally more expensive but enable relative pose estimation. For visual place recognition, local feature methods like SuperGlue~\cite{Sarlin2020SuperGlue:Networks} and LightGlue~\cite{Lindenberger2023LightGlue:Speed} are often repurposed for this refinement stage.

\subsubsection{Main algorithmic approaches}
\label{sec:algorithmicapproaches}
The evolution of place recognition training has followed a clear progression driven by practical limitations and theoretical insights. Early deep learning approaches adopted triplet loss, where the networks are trained to embed similar places closer together while pushing dissimilar places apart~\cite{Arandjelovic2018NetVLAD:Recognition}. However, most triplets become ``easy'' after initial training, motivating the development of listwise approaches~\cite{Revaud2019LearningLoss} and more sophisticated contrastive methods like generalized contrastive loss~\cite{Leyva-Vallina2023Data-EfficientSupervision} that better capture ranking relationships inherent in place recognition. Similarly, mining strategies are becoming more efficient by exploiting geometric structure in the embedding space~\cite{Izquierdo2024CloseRecognition}. An in-depth discussion of these loss functions can be found in~\cite{Berton2025AllModels}.

Some recent work has reframed place recognition as a classification task rather than retrieval~\cite{Berton2022RethinkingApplications}, representing a paradigm shift from ``general-purpose'' systems to environment-specific ones. This approach has gained popularity in resource-constrained scenarios, particularly with spiking neural networks~\cite{Hussaini2024ApplicationsRecognition,Hines2025ALocalization}, and highlights a fundamental question: can we create universal systems that work well everywhere, or is some specificity required? The preponderance of paper titles with ``any'', ``mega'' and ``ultimate'' in the title indicate at least a thematic shift towards this goal.

A significant recent development involves adapting general-purpose vision foundation models for place recognition~\cite{Izquierdo2024OptimalRecognition,Lu2024SuperVLAD:Recognition}, achieving superior performance across diverse environments. This success suggests these models may discover universal visual patterns for spatial understanding, as evidenced by their ability to excel at robotic place recognition despite being trained on general internet imagery with minimal task-specific modifications. Such universality raises intriguing questions about whether these representations transcend specific robotic applications and potentially capture fundamental spatial encoding principles that might align with biological systems, though this remains speculative.

Interestingly, there remains a notable divide between loop closure approaches used in SLAM and dedicated VPR methods. Many SLAM algorithms still rely on simple Bag-of-Words approaches~\cite{Mur-Artal2015ORB-SLAM:System,Fontan2024AnyFeature-VSLAM:SLAM,Tsintotas2022TheDetection,Tsintotas2021Modest-vocabularyWords}, while VPR research has moved toward more sophisticated deep learning methods. This gap represents both a challenge and an opportunity for better integration of modern VPR techniques into practical robotic systems. Anecdotally, researchers in the field often note with interest that many of the visual SLAM systems being used in labs around the world date back a decade.

\subsubsection{Modalities}
\label{sec:modalities}
While foundational concepts apply across sensor modalities, the robotics community has historically maintained distinct research threads for different sensing approaches. Visual and LiDAR place recognition represent the most common modalities, yet have developed largely in parallel with limited cross-pollination.

Even within single modalities, fragmentation exists. For example, visual place recognition emphasizes image retrieval, landmark recognition focuses on object identification, and 6-degrees-of-freedom localization prioritizes precise pose estimation—using different datasets and metrics. This fragmentation has led to duplicated efforts, though recent work like MegaLoc~\cite{Berton2025MegaLoc:All} demonstrates that bridging these areas can improve overall performance.

Beyond vision and LiDAR, specialized sensors offer unique advantages. Radar provides improved all-weather operation when rain, fog, and snow degrade visual systems~\cite{Gadd2024Open-RadVLAD:Recognition}. Event cameras suit high-speed, low-latency applications~\cite{Gallego2020Event-basedSurvey}. Underwater environments necessitate sonar-based recognition due to limited light penetration~\cite{Kim2023RobustEnvironments}. These challenges reveal that no single sensor modality provides universal place recognition capabilities.

Multi-modal fusion has demonstrated improvements over single-modality approaches, but the benefits are not universal~\cite{Li2025VXP:Recognition}. Methods like AdaFusion~\cite{Lai2022AdaFusion:Recognition} adaptively weight sensor contributions based on conditions, emphasizing visual features in well-lit areas while relying more heavily on geometric information in challenging lighting. Bird's-eye-view approaches, commonly used in autonomous vehicles, project multi-modal data into common overhead representations to provide interpretable spatial relationships that align with human understanding of navigation tasks~\cite{Fu2024CRPlace:Recognition}. However, fusion introduces challenges: sensor miscalibration can degrade performance below single-sensor baselines, and computational overhead may prevent real-time operation on resource-constrained platforms~\cite{Hayoun2024PhysicsLearning,Lanegger2024ToRobots}.

\subsubsection{Architectural Considerations}
\label{sec:architectureconsiderations}
The distinction between global place recognition and prior-constrained matching represents a fundamental architectural decision that profoundly influences system capabilities and computational requirements. Global matching approaches treat each query as a search over the entire place database without geometric or temporal priors, maximizing robustness in kidnapped robot scenarios and enabling detection of long-range loop closures. However, global matching faces scalability challenges as database size grows and becomes susceptible to perceptual aliasing in repetitive environments. From a practical perspective, global matching is also wholly unnecessary in many applications where an approximate spatial prior -- for example periodic access to GNSS -- can reasonably be assumed to be available, and where the occasional ``global relocalization'' requirement can be dealt with by exception, rather than as a continual process.

Another consideration is whether place recognition operates as a modular component or within an end-to-end system. Modular architectures like ORB-SLAM~\cite{Mur-Artal2015ORB-SLAM:System} treat place recognition as a discrete component with well-defined interfaces, enabling easier debugging and algorithm swapping without modifying the broader system architecture -- although there are still limits to the type and class of place recognition system that can be easily swapped in and out. End-to-end approaches~\cite{Teed2021DROID-SLAM:Cameras} argue that modularity imposes constraints preventing optimal performance, though such systems can struggle to generalize beyond training distributions, which is problematic for safety-critical robotic applications~\cite{Sunderhauf2018TheRobotics}.

\subsubsection{Platform differences}
\label{sec:platformdifferences}
The physical and operational characteristics of different robotic platforms impose fundamental constraints on place recognition system design beyond sensor selection. Underwater vehicles operate in environments where visual methods face severe limitations due to water turbidity and dynamic lighting, necessitating sonar-based approaches~\cite{Kim2023RobustEnvironments}. Aerial platforms face computational and payload constraints due to Size, Weight, and Power (SWaP) limitations, forcing trade-offs between algorithm sophistication and real-time performance~\cite{Ferrarini2019VisualDescriptors}. They also experience dramatic viewpoint variations with 6-degrees-of-freedom motion, generating more extreme perspective changes than ground-based systems~\cite{Maffra2018Viewpoint-TolerantNavigation}. Ground robots represent the most mature domain, but still exhibit platform-specific variations. For example, different drive systems affect motion priors and practical sensor configurations~\cite{Cadena2016PastAge}. One criticism of the place recognition research field to date has been an overly-heavy focus on solving datasets ``on rails'' -- that is, from vehicles or trains repeatedly traversing very similar routes with little to no viewpoint variation.

\subsubsection{Compute and storage constraints}
\label{sec:constraints}
Most place recognition techniques scale linearly with database size, creating trade-offs between descriptor dimensionality, database size, and achievable accuracy. Recent work maintains high recall with lower dimensions~\cite{Berton2022RethinkingApplications} or uses alternative data structures~\cite{Schubert2021FastRecognition} to reduce search complexity.

These optimizations must balance size, weight, and power (SWaP) constraints that vary dramatically across robotic platforms. Memory-constrained systems may require quantization~\cite{Ferrarini2022BinaryEnvironments} or hashing approaches~\cite{Li2024PyramidRecognition} that sacrifice accuracy for reduced storage. Power-limited platforms drive interest in neuromorphic approaches like spiking neural networks~\cite{Hussaini2024ApplicationsRecognition,Hines2025ALocalization} that promise significant energy reductions. 

There is also a clear separation needed between the amount of data and compute required to train a model, versus what is required at deployment (inference) time. Training environments can often be well supplied with compute resources, although this is not so feasible if a robot platform must train or re-train its systems ``in the field''.

The weight penalty of additional sensors and compute hardware creates cascading effects: each extra sensor requires mounting hardware, cabling, and processing units that reduce payload capacity for other mission-critical components. These SWaP constraints force system designers to make explicit trade-offs between place recognition capability and other system requirements, often leading to sensor selection decisions that prioritize passive sensors (cameras) over active ones (LiDAR) despite potential performance advantages, or choosing lower-resolution sensors to meet power budgets.

\subsection{What Can We Learn From This System?}
\label{sec:whatcanwelearn}
Despite significant advances, several fundamental challenges remain in place recognition for robotics. Nighttime conditions and large viewpoint variations continue to pose significant difficulties for VPR systems. Many approaches require dense and well-curated reference maps, which can be expensive to maintain and update. Paradoxically, many datasets suffer from the very problem that VPR tries to solve—ground truth annotations are often obtained via GPS, which itself is not trustworthy or robust in many instances where place recognition is most needed. This creates a circular dependency: researchers develop place recognition systems to work in GPS-denied environments, but then evaluate these systems using datasets labeled with GPS coordinates that are unreliable in precisely those challenging conditions. The result is that benchmark datasets may contain incorrect ground truth labels, leading to systems that appear to fail in some instances when they are actually correct. Works like Patch-NetVLAD~\cite{Hausler2021Patch-NetVLAD:Recognition} and MegaLoc~\cite{Berton2025MegaLoc:All} have exposed many such instances in standard benchmarks, revealing systematic errors in datasets that the community has relied upon for years.

Many principles in robotics place recognition apply even for extreme scenarios such as aerial-to-ground matching~\cite{Guan2023CrossLoc3D:Recognition} and space-to-ground localization~\cite{Berton2024EarthLoc:Space}. This suggests that fundamental representation learning principles may be more universal than initially believed, extending beyond traditional terrestrial robotics applications.

The evolution of robotics place recognition suggests several important directions for the field. The successful adaptation of foundation models indicates that future advances may come from better leveraging general-purpose representations rather than developing increasingly specialized architectures (see Section~\ref{sec:algorithmicapproaches}). The growing emphasis on multi-modal integration suggests that robust place recognition will rely on sensor fusion rather than relying on any single modality (Section~\ref{sec:modalities}).

This unification trend extends beyond pure research to practical system integration. Modern autonomous systems increasingly require capabilities from multiple communities: coarse place recognition for initial localization, fine-grained landmark detection for precision navigation, and robust 6-degrees-of-freedom pose estimation for manipulation tasks. Frameworks that seamlessly integrate these capabilities enable more capable and robust autonomous systems while reducing the engineering effort required for system integration.

\section{PLACE RECOGNITION IN ANIMALS}
\label{sec:animal_pr}
Organisms -- ranging from primitive multicelled to modern insects and large mammals -- have long had the need to navigate their habitat and recognize where they are. Unlike artificially created systems, animal behaviors related to recognizing places can only be inferred through observation, and carefully designed experimental paradigms. Animals cannot generally be explicitly asked to relate how they recognize a place, like humans covered in Section~\ref{sec:human_pr}, but the detailed neuronal activity in their brains can be examined in detail, to an extent not yet possible in humans. Two of the most famous examples of place-recognition-related characteristics in the animal kingdom are a) the clear neuronal substrates found in their brains underlying mapping and navigation: a momentous discovery that led to the 2014 Nobel Prize in Medicine and Physiology, and b) the amazing navigational capabilities of animals and insects, as well as the diversity of environments in which this capability is demonstrated around the world.

\subsection{What Is a ``Place'' in This System?}

In animals, the concept of a ``place'' is not directly accessible in an objective or universal sense. Instead, we infer its structure and function by observing behavior, neural activity, and ecological outcomes. What counts or has meaning as a place varies dramatically across species and situations – tied to a combination of sensory cues, functional goals, and evolved representations. As such, place is not a fixed coordinate but an emergent construct shaped by perception, memory, and task relevance.

\begin{figure}
\includegraphics[width=4in]{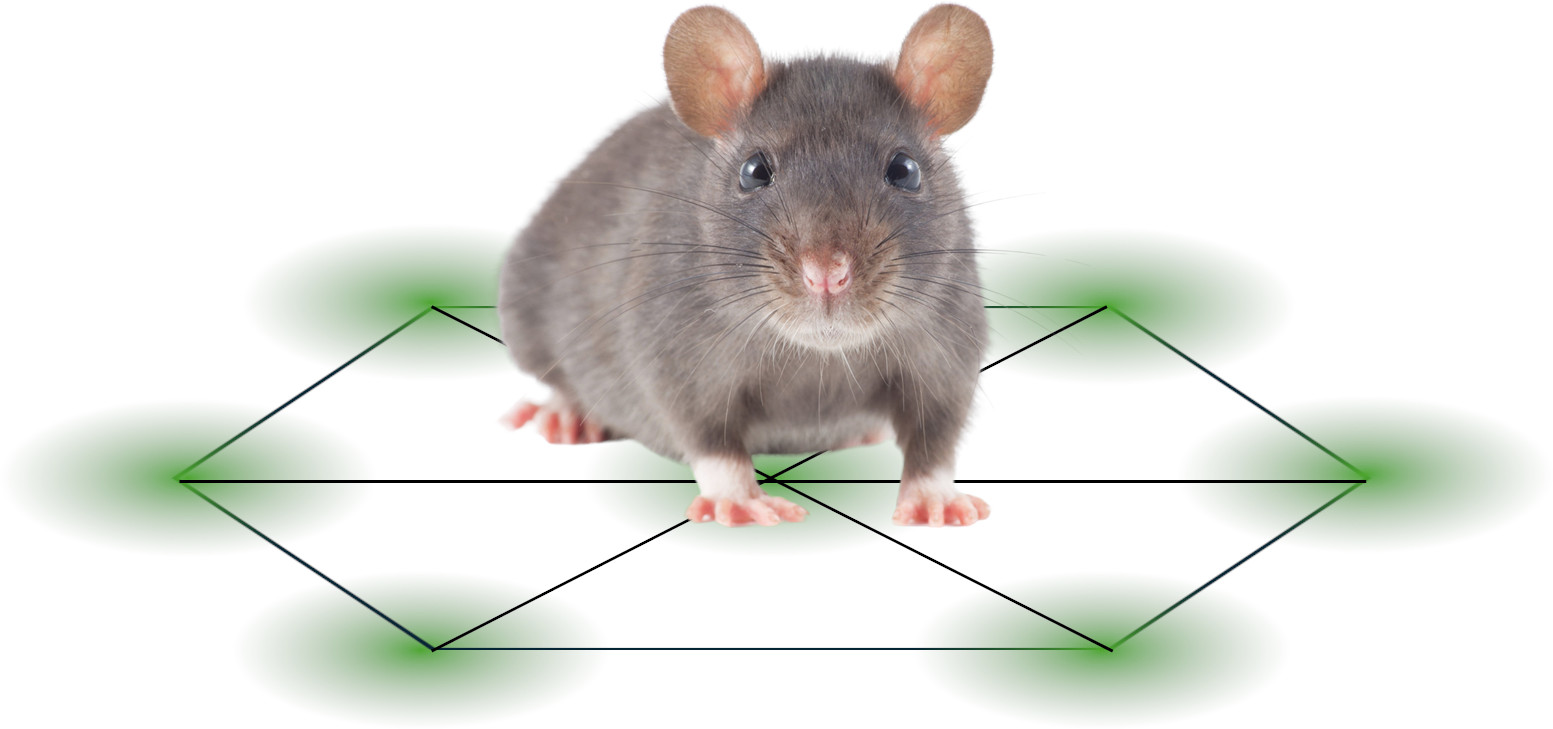}
\caption{The study of spatial representation in the mammalian brain has revolved around the discovery in the 1970s onwards of spatially responsive cells including place, head-direction and grid cells, as well as a range of related cell types including boundary or barrier-sensitive types. These cells have both overlapping and somewhat counterintuitive properties to artificial place recognition systems: for example grid cells code for multiple locations arranged over a tessellating triangular grid, and over a number of different scales.}
\label{fig3}
\end{figure}

Allocentric representation in rodents reveals how ``place'' emerges from specific neural firing patterns. Hippocampal place cells fire in relation to specific spatial positions in an environment, creating discrete spatial fields that define bounded locations within an environment \cite{okeefe1976}. This representation of place is world-relative rather than self-centered, forming internal cognitive maps that persist across sessions (Figure~\ref{fig3}). Grid cells in the entorhinal cortex provide metric coordinates through hexagonal firing patterns, while border cells and head direction cells add geometric boundaries and orientation \cite{hafting2005}. These cells are modulated by environmental geometry, novelty, and reward – highlighting that place in rodents is not only geometric but context-sensitive. It should also be noted that the extent to which explicit spatial representations are encoded in the brain has been the subject of a long-running debate in the neuroscience community \cite{eichenbaum1999hippocampus}, and is a point of timely relevance given the increase in new robot and agent navigation techniques with reduced reliance on traditional maps, a point later discussed in the Discussion.

\begin{marginnote}[]
\entry{Allocentric Representation}{A spatial encoding that defines locations relative to external landmarks or a map, independent of the observer’s viewpoint.}
\entry{Cognitive Map}{A mental representation of spatial relationships between objects and places, supporting flexible navigation.}
\entry{Place Cell}{A neuron in the hippocampus that activates when an animal occupies a specific location in its environment.}
\entry{Grid Cell}{A neuron in the entorhinal cortex that fires at multiple locations forming a hexagonal grid pattern across the environment.}
\end{marginnote}

Different species operate across dramatically different spatial scales, fundamentally shaping their place concepts. Insects work within meter-scale environments where individual landmarks define places. In ground-based insects like ants, the very idea of a ``place'' can be reconstructed from compound sensory panoramas. Desert ants combine celestial compass data with visual landmarks and internal path integration to return to a precise location in feature-poor terrain \cite{wehner2003}. Dung beetles operate on even smaller spatial scales, memorizing celestial orientations for dung pile locations  \cite{dacke2003}. In bees and wasps, place is influenced by visual and functional context – a flower patch or hive entrance is encoded via panoramic snapshots and their relative alignment during flight \cite{cheng1998}. Bats – uniquely, the only flying mammals – exhibit three-dimensional place coding in the hippocampus, with cells responding to both horizontal and vertical dimensions of volumetric space \cite{yartsev2011}. This represents a fundamental conceptual shift from surface-based navigation to true volumetric spatial encoding, where ``place'' becomes a position within a 3D coordinate system rather than a location on a 2D plane.

\begin{marginnote}[]
\entry{Path Integration}{A navigation strategy where an animal updates its position relative to a starting point by integrating cues from self-motion.}
\entry{Celestial Compass}{A navigation system that uses the sun, moon, stars, or polarization patterns in the sky to maintain heading direction.}
\end{marginnote}

\begin{textbox}[ht]
\subsection{Monarch Butterfly Navigation}
Monarch butterflies (\textit{Danaus plexippus}) undertake a remarkable annual migration from North America to central Mexico, covering thousands of kilometers across multiple generations and with high directional fidelity \cite{reppert2010navigational, reppert2018}. Navigation is achieved through a time-compensated sun compass mechanism that integrates the sun’s position with circadian timing. The circadian clock responsible for this compensation is located in the antennae rather than solely in the brain, allowing the butterflies to adjust for the sun's movement across the sky over the day.

The central complex in the monarch brain is believed to act as the integrative center for processing celestial compass information. Neurons in this region receive sun angle data that is modulated by circadian inputs, enabling sustained directional flight. This compass system is notably robust to environmental perturbations: the antennal clocks are light-entrainable even when isolated from brain clocks, demonstrating adaptive redundancy.

The monarch’s navigation system illustrates decentralized computation in biological organisms, where key functionalities like timing and orientation are distributed beyond the central brain. This biological strategy offers promising inspiration for building distributed, resilient navigation mechanisms in autonomous systems. Notably, there is little evidence of a sophisticated place recognition system, at least by conventional robotics standards: their ``place'' is defined by large-scale environmental signatures rather than fine-grained spatial landmarks. 
\end{textbox}

Whales, seals, and turtles migrate across vast and seemingly featureless oceanic environments, with few reliable visual cues. In such species, place may be encoded via geomagnetic maps, water salinity gradients, or seafloor bathymetry \cite{lohmann2008}. Some whale species are thought to memorize undersea landmarks and acoustic profiles of ocean basins – potentially building a ``place'' from echoic and geophysical features \cite{allen2013}. Large-brained terrestrial animals, such as elephants, rely on episodic-like memory, social knowledge transfer, and long-term spatial familiarity. Migratory routes and waterhole locations are passed on generationally, and place knowledge may encompass temporal rhythms such as seasonal rainfall or fruiting patterns \cite{presotto2019}. Place in this instance likely has a strong temporal affordance as well.

The navigational feats of migratory birds and monarch butterflies (see Sidebar) further blur the boundaries of what ``place'' is taken to mean. Migratory songbirds, like warblers and swallows, calibrate their internal compass systems using both solar and geomagnetic cues – and can return to specific breeding or stopover sites with meter-level precision \cite{wiltschko2006}. These places exist as magnetic and celestial coordinate targets that persist across continental distances.

Across all these species, ``place'', at least as it has been studied behaviorally and neurally, is rarely defined only in absolute spatial terms. Functionally, place is likely a product of what matters – shelter, food, mates, threats – combined with what is perceivable and learnable given the animal’s morphology and neural architecture. Whether based on snapshots, gradients, cues, or fields, animal place representation reflects an adaptive and task-embedded abstraction – one that perhaps differs from many robotic or human concepts of space. Place, in short, is not universal – it is experienced.

\subsection{How Is Place Recognized?}

\begin{marginnote}[]
\entry{Remapping}{The dynamic reorganization of spatial firing patterns in place or grid cells in response to environmental changes.}
\end{marginnote}

Place recognition in animals is deeply multimodal, presumably shaped by evolution to prioritize cues that are robust, repeatable, and salient within each species’ ecological niche. Rodents demonstrate sophisticated cue integration where visual, olfactory, tactile, and proprioceptive inputs are dynamically weighted based on reliability and context. A remarkable feature of rodent place recognition is neural remapping – when environments change, place cells reorganize their firing patterns \cite{okeefe1976,hafting2005}. Place is therefore not a fixed label but is constantly recalibrated using environmental boundaries, sensory features, and learned associations. Bats, navigating in three dimensions, are distinguished from other mammals by their reliance on range-sensitive echolocation. Their place recognition involves temporally dynamic sonar sampling, combined with inertial and visual information. Hippocampal recordings in bats show tuning to both horizontal and vertical positions, underscoring their adaptation to 3D volumetric navigation \cite{yartsev2013}.
Ants perform dead reckoning using step counting (odometry) and solar compass cues, refining their trajectories with local landmarks even in challenging terrain \cite{wehner2003}. Bees capture panoramic views of key locations like nest entrances and flowers, using retinotopic alignment to re-identify places during return trips \cite{collett1998}.

Marine animals such as sea turtles and whales rely on large-scale geophysical cues. Sea turtles imprint on the geomagnetic signature of their birth beach and return decades later using this magnetic map \cite{lohmann2008}. Whales are hypothesized to use bathymetry, temperature gradients, and acoustic signatures of undersea geography for navigation, although empirical evidence remains limited \cite{allen2013}.

Elephants recognize migratory corridors, waterholes, and feeding grounds using long-term spatial memory, olfactory sensing, and knowledge transmitted through social groups \cite{hoerner2023long}. These patterns often reflect decades of collective memory and decision-making, activated through seasonal or contextual triggers rather than immediate sensory cues.

\begin{marginnote}[]
\entry{Magnetoreception}{The ability of animals to detect the Earth’s magnetic field for orientation and long-distance navigation.}
\end{marginnote}

Migratory animals process multiple environmental cues simultaneously to achieve recognition across hemispheric distances. Monarch butterflies rely on a circadian-solar compass, encoded in their antennae and compound eyes, to travel thousands of kilometres \cite{zhu2008}. Migratory birds likely utilize magnetoreception, modulated by celestial cues and the Earth's magnetic field parameters, to locate specific regions across hemispheric distances \cite{wiltschko2005}. Experimental displacements show that these systems are redundant – birds can compensate when one modality is blocked or misleading.

Across species, place recognition emerges from an integration of perception, memory, and prediction. The weighting of cues adapts dynamically based on environmental stability or ambiguity. Animals deploy fallback strategies – switching from path integration to landmark use or systematic search \cite{narendra2008search}, for example – in the face of novelty or noise. These systems are not modular but reflect deeply embodied sensorimotor loops, where ``place'' is an active construct refined by feedback. The remarkable diversity of solutions – from bee vision to whale-scale sonar maps – reveals that place recognition in nature is not about map-building in a classical sense, but about acting effectively and robustly within spatial contexts. Each system reflects a tailored answer to the fundamental question: How do I know where I am, and what can I do here?

\subsection{What Can We Learn From This System?}

Animal navigation systems offer a treasure trove of insights into the nature of place, its representation, and its recognition. Although these systems are deeply embedded in species-specific biology and ecological context, they exhibit recurring principles that provide valuable lessons for artificial agents – especially in perception, autonomy, and long-term robustness.

One striking takeaway is the task-oriented, affordance-based nature of place. For animals, place is rarely an abstract coordinate in a geometric map. Instead, it is tied to functional outcomes – locating food, ensuring safety, enabling social behavior, or facilitating reproduction – and is represented in ways that support effective action rather than passive description \cite{zoladek1978}. This perspective shifts emphasis from traditional map-building toward goal-directed spatial cognition. It also highlights how representations evolve under selective pressures: what matters is not so much precision but utility.

A second insight lies in multimodal robustness and redundancy. Animal navigation leverages overlapping sensory systems – visual panoramas, celestial references, olfactory trails, magnetic fields, proprioception – dynamically weighted by reliability and context. Ants, for example, switch strategies when displaced mid-route \cite{wehner2003}. Migratory birds recalibrate their magnetoreception with solar or stellar inputs \cite{wiltschko2005}. Animals also learn place representations through embodied interaction. Spatial knowledge is acquired via exploration, constrained and shaped by the body and behavior. Bees memorize views through characteristic looping learning flights \cite{collett2018}. Rodents explore new environments by actively sampling spatial cues with head movements and whisking \cite{thompson2018}. This suggests that place learning involves sensorimotor engagement – more than passive storage, it is hypothesis testing through movement.

Place representations in animals are plastic yet organized. Neural systems such as the hippocampus exhibit remapping when environments, cues, or rewards change – enabling flexibility without losing structure \cite{agmon2020}. Importantly, these changes preserve spatial coherence, often reusing structure across contexts – as in grid cell realignment or rescaling. This supports generalization while maintaining internal consistency.

Biological systems also show that place recognition, and higher-level navigational feats, do not need to rely on metric maps. Insects and birds navigate effectively using sparse, view-based, or field-based representations that are aligned to local cues and sensor-specific snapshots \cite{mangan2012}. These models tolerate occlusion, operate with limited computation, and forgo the need for continuous localization – a valuable lesson for designing scalable, efficient systems. Context, memory, and motivation further shape animal navigation. The relevance of a place can shift with time, season, or internal state. Elephants recall waterholes and migration routes over decades, guided by long-term memory and social learning \cite{hoerner2023long}. 

Finally, animals provide examples of error handling and compensation. Disoriented birds recalibrate after displacement. Bees disrupted by magnetic or visual anomalies eventually reorient using backup strategies. These systems exhibit introspective features – sensing when something is wrong and adjusting accordingly – a capability absent from many artificial systems today.

In sum, animal navigation teaches us that ``place'' is a dynamic, multimodal, task-embedded concept. It is shaped by sensory experience, grounded in embodied action, structured by learning, and defined by adaptive goals. Although biological mechanisms differ from engineered systems, their observed properties – robustness, embodiment, plasticity, and self-monitoring – offer compelling templates for the development of more capable and resilient artificial place recognition systems.

\section{PLACE RECOGNITION IN HUMANS}
\label{sec:human_pr}

Whilst humans are animals, a separate treatment is warranted because of the ability to interrogate and explain place recognition and navigation in a way that is not possible with animals, and because of the combination of ``natural'' navigational capabilities with modern constructs of navigation imposed on many humans by a highly urbanized and digital world.

\subsection{What Is a ``Place'' in This System?}

In humans, ``place'' is a richly constructed concept – perceptual, cognitive, emotional, social, and symbolic. While it shares foundational elements with other animals, human notions of place extend far beyond immediate sensorimotor contexts. A place has meaning, shaped by memory, language, culture, and intent.

At the neural level, humans exhibit place-specific activity in brain regions homologous to other mammals. Hippocampal place cells have been identified through direct recordings, and spatial view cells – which respond to where the person is looking within an environment – have been observed in primates \cite{rolls1999}. Grid-like representations have been inferred in the entorhinal cortex using fMRI during virtual navigation, suggesting that humans maintain an internal coordinate system akin to that found in rodents \cite{horner2016}. 

Humans regularly describe, understand, and navigate space using abstract or semantic categories: home, danger zone, sacred ground, public space. These functional, emotional, and social labels are not reducible to physical coordinates. A classroom and a courtroom may have similar geometry but very different meanings. Such schemas are part of embodied and situated cognition – a view that holds human cognition is tightly linked to action, perception, and environmental context \cite{dourish2001}.

From childhood, humans begin associating places with people, emotions, and events. Concepts like ``my room,'' ``grandma’s house,'' or ``the scary alley'' form through a mix of direct experience, social transmission, and storytelling. Developmental research shows that children form spatial categories early, informed by both perceptual clustering and linguistic input \cite{newcombe2000}, and update these categories with new experience. Crucially, these memories are not just attached to places – they are often place-structured. The hippocampus, which supports episodic memory, shows place-specific reactivation when people recall events, suggesting that remembered events are indexed by spatial context \cite{staresina2012}. This may be why autobiographical recall is so often tied to locations – such as remembering where you were when you heard a piece of news.

\begin{marginnote}[]
\entry{Episodic Memory}{Memory of specific events tied to particular spatial and temporal contexts, often used to structure place-based recall.}
\end{marginnote}

\begin{figure}
\includegraphics[width=\textwidth]{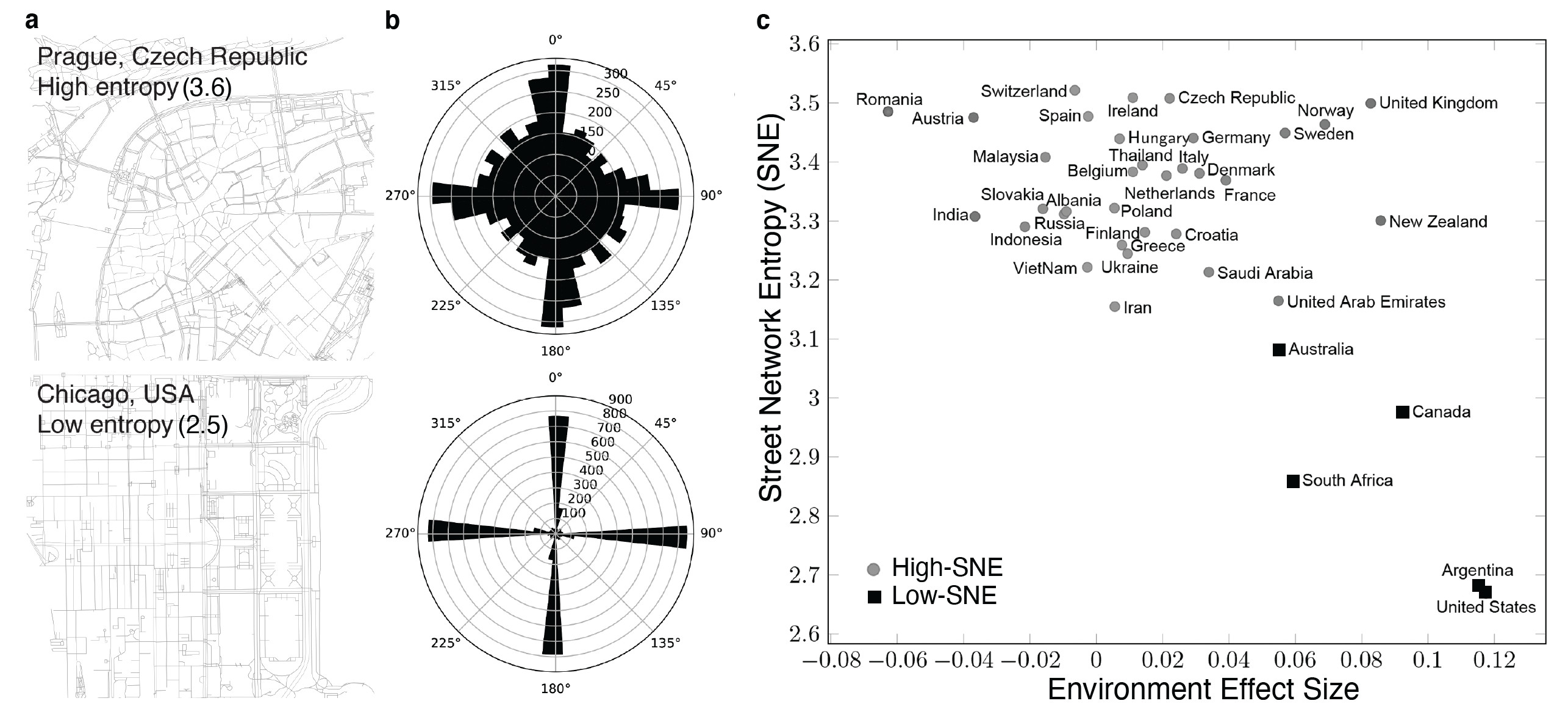}
\caption{Street Network Entropy (SNE) and its relationship to upbringing environment across 38 countries \cite{Coutrot2020}. (a) Examples of cities with low SNE (Chicago, USA) and high SNE (Prague, Czech Republic), reflecting different street layout patterns. (b) Street orientation distributions in each city, divided into 36 directional bins. (c) Average SNE plotted against whether participants were raised in urban or non-urban environments. Positive values mean that people raised outside cities performed better on spatial tasks than those raised in cities. Shapes indicate two country groups identified using $k$-means clustering: squares for low-SNE countries and circles for high-SNE countries. Reproduced with permission from Springer Nature.}
\label{city_navigation}
\end{figure}

Recent research \cite{Coutrot2020} shows that the entropy of urban street networks---the degree of disorder in street orientations---correlates with human spatial navigation abilities. This has significant implications for how we design and evaluate artificial place recognition and localization systems. The study found that people raised in high-entropy cities (e.g., with irregular, winding streets) tend to perform worse on structured navigation tasks but better in unstructured navigation tasks when compared with those raised in low-entropy, grid-like cities (Figure \ref{city_navigation}). 

Humans also develop strong affective ties to place. Emotional salience can transform a neutral setting into a personally meaningful place – from a childhood playground to a battlefield. Neuroimaging research shows distinct activation patterns when people imagine familiar, emotionally significant places versus generic or unfamiliar ones \cite{sugiura2005}. Environmental psychology builds on this with concepts like place attachment and place identity, emphasizing how places become intertwined with personal and collective identity \cite{manzo2020}.

\begin{marginnote}[]
\entry{Place Attachment}{The emotional and identity-based bond between a person and a specific location.}
\end{marginnote}

Moreover, humans can experience place without physical presence. The hippocampus and associated medial temporal structures activate during mental navigation of both real and imagined environments, as shown in studies of episodic simulation and narrative construction \cite{schacter2008}. Literature, cinema, and digital worlds can evoke powerful senses of place – suggesting that spatial experience in humans can be decoupled from physical embodiment.

Cross-cultural studies further highlight the diversity of human place concepts. Some languages encode absolute spatial orientation (e.g., cardinal directions) in everyday speech, affecting how individuals track location and remember paths. In these cultures, place is understood relationally – as a set of fixed bearings rather than egocentric representations \cite{levinson1996relativity}. Cultural tools and habits thus reshape how place is experienced and represented.

\begin{marginnote}[]
\entry{Egocentric Representation}{A spatial encoding based on the observer’s current position and orientation, typically in first-person view.}
\end{marginnote}

In sum, place in humans is layered and multifaceted – grounded in perceptual reality, yet extended by narrative, social meaning, and imagination. It can be physical, remembered, imagined, or symbolic. Recognizing this breadth is essential for understanding how humans navigate, value, and design spaces – and for translating those insights into artificial systems that seek to interact with human environments.

\subsection{How Is Place Recognized?}

Humans recognize place using a flexible, multi-layered integration of sensory cues, semantic knowledge, spatial memory, and social context. Unlike many animals, humans can recognize a place both through direct perception and through symbolic, linguistic, or associative channels – enabling place recognition even in the absence of sensory similarity.

Sensory modalities play a foundational role in human place recognition. Visual landmarks – such as buildings, signs, terrain, or architectural layout – are primary cues, but auditory (e.g., traffic, birdsong), olfactory (e.g., ocean breeze, fresh-cut grass), and tactile (e.g., ground texture) inputs also contribute. These cues are fused into scene schemas – structured expectations of what a place should contain and how it should feel \cite{brewer1981}.

At the neural level, visual place recognition recruits a distributed network of brain regions. The parahippocampal place area (PPA) responds selectively to scene layouts and environmental geometry \cite{epstein1998}, while the retrosplenial cortex (RSC) supports transformations between egocentric and allocentric representations \cite{vann2009}. The hippocampus encodes place via map-like representations and supports memory-guided navigation \cite{tolman1948,okeefe1978}. Neurologically, topographical disorientation syndromes demonstrate dissociable failures – some patients cannot recognize landmarks (landmark agnosia), others cannot link them to spatial context (heading disorientation), or cannot orient despite intact perception (heading loss) \cite{aguirre1999}.

Crucially, place recognition in humans is robust to appearance change. We can recognize a street at night, a familiar path in reverse, or a remodeled room based on spatial layout, semantic cues, or episodic context. This invariance suggests that humans rely on multi-modal, abstracted representations – not just raw pixel-level similarity. Eye-tracking and recall studies show that humans fixate on informative, distinctive landmarks rather than photometric details \cite{steck2000}.

Memory plays a central role. Place recognition is often a form of episodic recall – recalling a situation, event, or experience tied to a specific location \cite{tulving2002}. Hippocampal activity during recall mirrors that seen during navigation or viewing, suggesting reinstatement of spatial context. In older adults or those with hippocampal damage, deficits in place recognition often parallel episodic memory impairments \cite{burgess2002}.

\begin{marginnote}[]
\entry{Episodic Recall}{The act of mentally reliving past experiences, often tied to spatial and contextual cues from specific locations.}
\end{marginnote}

Place recognition also engages semantic memory and cultural knowledge. People can identify places they’ve never visited through media exposure or narrative – e.g., recognizing Times Square or the Eiffel Tower. This symbolic recognition depends less on metric cues and more on symbolic familiarity and social learning \cite{ryan2009}. Humans also recognize places through social and emotional cues. A street may be identified by who lives there, or a café remembered by a conversation.

\begin{marginnote}[]
\entry{Symbolic Recognition}{Identifying places using abstract cues such as names, narratives, or schematic representations rather than direct sensory input.}
\end{marginnote}

\subsection{What Can We Learn From This System?}

Human place recognition systems exhibit several properties that differ from most engineered approaches. Rather than relying on pixel-level precision or geometric accuracy, humans use context-sensitive strategies that emphasize semantic relevance, multimodal cues, and prior experience. A single landmark can dominate perception, while many visual details may be ignored depending on task, lighting, or orientation \cite{steck2000}. Representations of place are typically structured and relational. Locations are organized into cognitive maps or semantic categories – for example, ``home'' or ``station'' – rather than being stored as isolated views \cite{o1979cognitive}. Recognition often involves reconstructing a scene from partial cues, guided by expectations or prior encounters \cite{tulving2002}. This likely enables robustness in dynamic or incomplete environments. Cross-modal recognition is also common – people identify places from language, maps, or conversation, not just visual input \cite{quamme2007}.

Human systems also display introspection. Uncertainty about location can lead to strategy shifts, such as slowing down, rechecking cues, or asking for help. This suggests value in recognition systems that can monitor confidence or detect mismatch. Finally, spatial understanding in humans is linked to action. Places are interpreted in terms of what they afford – what can be done there, or what typically occurs. This functional framing supports flexible behavior in novel or changing environments. These observations point to possible design principles for artificial systems: cue prioritization, structured representation, memory integration, cross-modal flexibility, and introspective mechanisms for handling uncertainty.

\section{DISCUSSION}

While each section has focused on the particular learnings from the ``platform'' being covered---robotics, animals and humans---here we discuss a number of key overarching concepts that can inform future work in robotic place recognition, as well as shape insights in both directions, between the robotics field on the one side, and  biological fields on the other.

\subsection{Functionally Informed Trade-offs}

Artificial visual place recognition systems have made substantial progress in recent years, propelled by advances in deep learning, increasingly diverse datasets, and the incorporation of foundation models. Performance benchmarks such as Recall@1 have improved steadily, and artificial systems now rival or exceed human-level perceptual matching in certain controlled scenarios. However, these gains often require growing computational complexity, extensive supervision, or narrow design assumptions. However, complexity is not always helpful, as shown by research like \cite{trivigno2024unreasonable}. As VPR systems approach saturation on existing benchmarks, further improvements increasingly involve trade-offs – in model size, inference speed, power requirements, and adaptability. This raises a critical question: are these trade-offs aligned with the intended downstream use of the system? In biological systems, place representations are likely not purely optimized for maximal geometric precision. Instead, they are shaped by ecological relevance, robustness under noise, and the capacity to support adaptive behavior. Artificial systems could similarly benefit from a shift in emphasis – from maximizing benchmark scores to optimizing for performance under real-world, task-specific constraints.

\subsection{The Disconnect Between Metrics and Functional Outcomes}
This leads to a second theme: the disconnect between how performance is measured in artificial systems versus biological ones. Standard metrics such as Recall@1 or localization error provide valuable proxies but may fail to capture the broader functional success of the system. In contrast, animals and humans are evaluated on goal completion, adaptability, and behavioral outcomes – whether a desert ant returns to its nest, or a rodent adjusts to a remapped maze. Neural recordings and behavioral assays provide richer context for interpreting place representation quality. Bridging this gap in artificial systems may involve the development of new benchmarks that reward goal-directed success, robustness under ambiguity, and graceful degradation – rather than only ranking systems on precision in idealized conditions. Learning-based approaches have already taken significant steps in this direction. Integrating confidence estimates, introspective uncertainty, and system-level feedback into evaluation pipelines could better align artificial metrics with natural ones.

\subsection{Modularity Versus Integration: Lessons from Biology}
A third consideration is system architecture. Traditional robotic systems favor modularity: VPR is one component in a larger SLAM or navigation stack, communicating via standardized interfaces. It is notable that it was much easier to divide the discussion of artificial systems up into the ``what is a place" and then ``how is place recognized" sections than the corresponding sections for animals and humans, where the distinction is less crisp. This modularity promotes transparency, debuggability, and reusability. In contrast, modern end-to-end systems seek tight integration – often jointly learning perception, localization, and planning functions. Each approach has merits. Modular designs are more interpretable and easier to adapt across domains. Integrated architectures may extract synergies between subsystems and enable greater overall efficiency. Biological systems once again suggest a middle ground: neural systems show both modular specialization and deeply integrated feedback loops, and both pre-wired configurations and online learning. For artificial VPR, the key may be not whether the system is modular or end-to-end, but whether it is co-designed with the functional context in mind. Place recognition systems should not exist in isolation – they must ultimately serve one or more higher level functions, whether they are navigation or decision-making processes. Architectures that permit joint optimization while maintaining some separability may best enable this goal.

\subsection{What Explicit Spatial Information is Actually Necessary?}

An interesting parallel has remained largely undiscussed between the biological research community studying navigation in natural organisms and artificial robotics community around \textit{what explicit spatial encoding is actually required for an organism to be functional}. In neuroscience, there has been a long-running debate over the extent to which explicit spatial encoding is stored in the brain \cite{eichenbaum1999hippocampus}. Whilst early robot SLAM research would, at least from a functional perspective, strongly suggest the need for explicit spatial encodings to enable navigation, the advent of deep learning and more recently so-called foundation models has enabled new approaches to robot mapping and navigation that at the very least challenge the level of dependence on rigid hierarchies of global and local metric or topological maps \cite{yokoyama2024vlfm}. The concept of mapless organisms that are nevertheless able to prosper in their ecosystem has been around for a long time in the biological community, but has mostly focused on relatively ``simple'' organisms: these new artificial approaches are starting to show that more complex and sophisticated systems may also be able to function without such a reliance on traditional pre-built maps of their environment.

\subsection{Continual Learning and Lifelong Adaptation}
Finally, the need for \emph{continual learning} is becoming more apparent. Whilst robot learning can be periodic or carefully controlled in some restricted application domains, for the vision of ubiquitous, useful robots throughout our work and personal spaces, continual learning is likely to be critical. Artificial VPR systems are often trained once, on a fixed dataset, and deployed without provisions for online adaptation. This sharply contrasts with biological systems, which learn continuously, update their spatial models from new experience, and recover from representational drift or forgetting. Lifelong operation in dynamic environments – whether urban streets, forests, or indoor spaces – demands systems that can integrate new observations, update representations without retraining from scratch, and selectively retain or discard memories. Current progress in self-supervised learning, memory consolidation, and model plasticity offers promising avenues, but significant challenges remain – including maintaining stability, avoiding catastrophic forgetting, and bounding the computational cost of online updates. Here, inspiration from natural memory systems – including mechanisms of synaptic consolidation, hippocampal replay, and context-gated recall – may offer useful inspiration. 

\subsection{Toward Functionally-Aligned Place Recognition}
Taken together, these themes, as well as others covered throughout this review, suggest a reframing of future VPR research around \emph{functional alignment}. Continued gains in performance will require not only algorithmic refinement, but also a broader recognition of the end-use context. Biological systems offer important lessons: performance is judged not by pixel-level similarity, but by action-oriented success; systems operate continuously, adapting to change; and architecture reflects both modular specialization and dynamic integration. Recasting VPR development with these principles in mind may unlock the next generation of artificial systems – ones that not only excel at Recall@1 benchmarks, but are also adaptive, robust, and ready for the real world.

\begin{summary}[SUMMARY POINTS]
\begin{enumerate}
\item The concept of a ``place'' varies across systems: in robotics, it is context-dependent based on downstream tasks; in animals it emerges from perception, memory, and ecological constraints; while in humans, it encompasses emotion, memory, culture, and social meaning – forming rich, multi-layered constructs grounded in both perception and symbolic representation.
\item Robotic place recognition enables localization without reliance on satellites, making it essential for robust navigation in environments where GNSS signals are unavailable or unreliable; it supports relocalization, loop closure, topological mapping, multi-robot collaboration, and long-term autonomy under changing environmental conditions.
\item Robotic techniques for defining places include keyframe selection, spatial segmentation, continuous similarity measures, and hierarchical representations that span multiple spatial scales.
\item Both animal and human place recognition systems are inherently multimodal and adaptive, though animals integrate cues like vision, magnetoreception, echolocation, and olfaction, while humans additionally use narrative, language, and cultural cues – allowing recognition of familiar or meaningful places even without direct perceptual similarity.
\item Place representations vary dramatically across biological systems -- from geometric cognitive maps in rodents and bats, to view-based recognition in bees and ants, to large-scale environmental signatures used by whales and migratory birds.
\item Neural representations of place in humans involve hippocampal and entorhinal systems similar to those in animals but extended by modules supporting memory, imagination, and semantic abstraction.
\item Human place recognition is deeply flexible and robust – integrating multimodal perception, episodic and semantic memory, and social context to support navigation, identity, and shared understanding.
\end{enumerate}
\end{summary}

\begin{issues}[FUTURE ISSUES]
\begin{enumerate}
\item Purpose-Aligned Design: Future VPR development should critically assess whether improvements in traditional performance metrics justify their complexity, with design choices driven by the practical demands of the end-use application.
\item Beyond Benchmark Metrics: New evaluation frameworks are needed that go beyond Recall@1 and static pose error, incorporating measures that are better predictors of eventual utility in a functional system, including worse case performance, variability and coverage measures. Metrics could also directly reflect real-world goals such as navigation success, safety, and robustness under dynamic environmental conditions.
\item Modular-Integrated Co-Design: Research should explore hybrid VPR architectures that balance modular transparency with the adaptability and performance of integrated, end-to-end learning.
\item Continual Learning: Long-term deployment requires VPR systems capable of online adaptation without catastrophic forgetting, supported by scalable, memory-efficient, and context-sensitive learning mechanisms.
\end{enumerate}
\end{issues}

\section*{DISCLOSURE STATEMENT}
The authors are not aware of any affiliations, memberships, funding, or financial holdings that might be perceived as affecting the objectivity of this review. 

\section*{ACKNOWLEDGMENTS}
This research was partially supported by the QUT Centre for Robotics, funding from ARC Laureate Fellowship FL210100156 to MM, and funding from ARC DECRA Fellowship DE240100149 to TF. The authors would like to thank all the colleagues and funding bodies who have supported, collaborated, and taken an interest in crossing these disciplinary boundaries.

\bibliographystyle{ar-style3}
\bibliography{references} 

\end{document}